\documentclass[runningheads]{llncs}
\usepackage{graphicx}

\usepackage{tikz}
\usepackage{comment}
\usepackage{amsmath,amssymb} 
\usepackage{color}

\usepackage{flushend}
\usepackage{enumitem}
\usepackage{bm}
\usepackage{mathtools}
\usepackage{algorithm}
\usepackage[noend]{algpseudocode}
\usepackage{multicol}
\usepackage{multirow}
\usepackage{pifont}
\usepackage{url}

\DeclareMathOperator*{\argmin}{argmin}
\newcommand{\cmark}{\ding{51}}%
\newcommand{\xmark}{\ding{55}}%

\begin{document}
\pagestyle{headings}
\mainmatter
\def\ECCVSubNumber{4748}  

\title{Learning Model-Blind Temporal Denoisers without Ground Truths} 

%
\author{Yanghao Li \inst{1} \and Bichuan Guo \inst{1} \and
Jiangtao Wen\inst{1}\and
Zhen Xia\inst{2}\and
Shan Liu\inst{2}\and
Yuxing Han\inst{3}}
\authorrunning{B. Guo et al.}
%
\institute{Tsinghua University, Beijing, China \and
Tencent Media Lab\and
Research Institute of Tsinghua University in Shenzhen, Shenzhen, China}
\maketitle

\begin{abstract}
 Denoisers trained with synthetic data often fail to 
  cope with the diversity of unknown noises, 
  giving way to methods that can adapt to existing noise
    without knowing its ground truth.
  Previous image-based method leads to noise overfitting if directly applied to video denoisers,
  and has inadequate temporal information management
  especially in terms of occlusion and lighting variation,
  which considerably hinders its denoising performance.
  In this paper, we propose a general framework for video denoising networks that successfully addresses these challenges.
  A novel twin sampler assembles training data by decoupling inputs from targets without altering semantics,
  which not only effectively solves the noise overfitting problem,
  but also generates better occlusion masks efficiently by checking optical flow consistency.
  An online denoising scheme and a warping loss regularizer are employed for better temporal alignment.
  Lighting variation is quantified based on the local similarity of aligned frames.
  Our method consistently outperforms the prior art by 0.6-3.2dB PSNR on multiple noises, datasets and network architectures.
  State-of-the-art results on reducing model-blind video noises are achieved.
  Extensive ablation studies are conducted to demonstrate the significance of each technical components.
\end{abstract}

\section{Introduction}
\begin{figure}[t]
\begin{center}
   \includegraphics[width=0.6\linewidth]{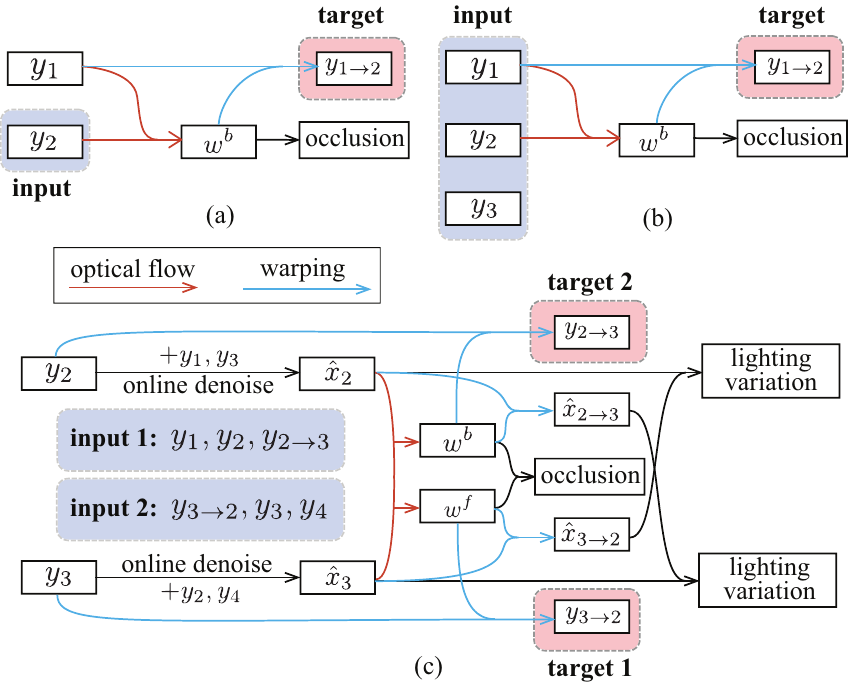}
\end{center}
   \caption{
   An overview of existing and our methods.
   Notations: 
   $y_i$ (noisy frames), 
   $w^f/w^b$ (forward/backward flow),
   $\hat{x}_i$ (denoised $y_i$),
   $f_{i\rightarrow j}$ ($f_i$ warped towards $f_j$).
   (a) Ehret et al. \cite{ehret2019model}. 
   Training inputs and targets are constructed by aligning adjacent frames.
   Occluded regions inferred from flow divergence are excluded.
   (b) The naive extension of \cite{ehret2019model} to video denoisers with multi-frame inputs. 
   Noise overfitting occurs due to pixels in $y_1$ appear both in inputs and targets.
   (c) Our method. By construction, any input and its target have no overlapping sources hence overfitting is avoided }
\vspace*{-5pt}
\label{fig:overview}
\end{figure}
Noise reduction is a crucial first step in video processing pipelines.
Despite the steady advancements in sensor technology, 
visible noises still occur when recording in low lighting conditions \cite{chen2018learning} 
or using mobile devices \cite{abdelhamed2018high}.
Therefore, effective denoisers are essential for 
achieving satisfactory results in downstream applications \cite{liu2019enhance,liu2018image}.

While there is a vast literature on reducing synthetic noises,
reducing noises without explicit models (i.e. model-blind) remains an essential and challenging problem.
On one hand,
until recently, most traditional \cite{dabov2007image,dong2012nonlocally,gu2014weighted,zoran2011learning} 
and data-driven methods \cite{chen2016trainable,zhang2017beyond,zhang2018ffdnet} 
assume additive white Gaussian noise (AWGN).
However, Pl\"{o}tz and Roth \cite{plotz2017benchmarking} showed that denoisers trained with synthetic AWGN often perform poorly on real data.
On the other hand,
creating training data by
synthesizing realistic noise \cite{claus2019videnn,guo2019toward} is non-trivial and prone to bias.
Alternatively one may estimate the ground truths (GT) of real photographs by adjusting exposure times \cite{anaya2018renoir,plotz2017benchmarking},
but this method is time-consuming and not viable for videos.
Also, 
such synthesis and training for all possible noises can be computationally prohibitive.
As a result,
self-adaptive methods that do not require explicit noise modeling or expensive GT datasets have drawn considerable attention lately \cite{batson2019noise2self,ehret2019joint,krull2019noise2void,lehtinen2018noise2noise}.

One of such methods is \textsc{frame2frame} \cite{ehret2019model} recently proposed by Ehret et al.,
which adapts an image denoiser to existing video noise.
It is built upon the \textsc{noise2noise} framework \cite{lehtinen2018noise2noise} which trains an image denoising network with noisy-noisy pairs (as opposed to normally using noisy-clean pairs).
Since the optimal weights under L1 loss is invariant to zero-median output noises,
\textsc{noise2noise} only requires the noisy pairs to have the same GT and independent median-preserving noise realizations.
By aligning video frames using optical flow,
the method \textsc{frame2frame} constructs such pairs from the to-be-denoised video (Fig.~\ref{fig:overview}(a)) and fine-tunes an image denoiser.
It can cope with a wide range of noises and is shown to outperform many image denoisers in frame-by-frame blind denoising.

Despite the success of \textsc{frame2frame} as a model-blind image denoiser,
several challenges remain that hinder its video denoising performance.
First, its performance is reliant on the optical flow quality.
However, many optical flow estimators only care about the flow error on clean image pairs.
This criterion does not prioritize warped results and can perform sub-optimally in aligning noisy frames.
Second, the key assumption of \textsc{noise2noise},
that noisy pairs have the same GT, 
is easily violated due to occlusion and lighting variation.
Finer correspondence management is needed.
Third, it cannot be directly applied to temporal denoising
where adjacent noisy frames are taken as inputs:
since these frames are also used in \textsc{frame2frame} to construct training targets, 
this {\it dual-presence} in both inputs and targets causes noise overfitting in static regions (Fig. \ref{fig:overview}(b)).

In this paper we propose a general framework for video denoising networks 
that successfully addresses all these challenges.
An overview of our method is shown in Fig.~\ref{fig:overview}(c).
The main contributions of this paper are:
\begin{itemize}
\item Temporal alignment is improved by employing an \textit{online} denoising scheme,
as well as a warping loss regularizer aiming to improve the content awareness of optical flow estimation networks.
\item Correspondence management is enhanced by aggregating two components: 
better occlusion masks are produced based on optical flow consistency;
lighting variation is measured based on local similarity.
\item We reveal the noise overfitting problem due to dual-presence suffered by the naive extension of image-based method, and propose a novel {\it twin sampler} that not only decouples inputs from targets to prevent noise overfitting,
but also enables better occlusion inference as a free by-product.
\end{itemize}

\section{Related work}
\noindent \textbf{Image and video denoising.}
Over the years, a myriad of image denoising algorithms have been proposed:
bilateral filters \cite{tomasi1998bilateral}, domain-transform \cite{moulin1999analysis,portilla2003image},  variational \cite{roth2005fields,rudin1992nonlinear} 
and patch-based methods \cite{dabov2007image,dong2012nonlocally,lebrun2013nonlocal,mairal2009non,zoran2011learning}.
The first attempt based on neural networks utilizes fully connected layers \cite{burger2012image}.
More recently, CNN-based methods have been proposed, 
including DnCNN \cite{zhang2017beyond}, FFDnet \cite{zhang2018ffdnet} and many others \cite{chen2018learning,mao2016image}.
These data-driven methods are commonly trained with noisy-clean pairs,
where noisy images are synthesized from clean images with a known noise model.
Our method is also data-driven, but tackles a more difficult case where the noise model is unknown.

Comparing to image denoising, the literature addressing video denoising is more limited.
VBM3D \cite{dabov2007video} extends the image-based BM3D \cite{dabov2007image}
by searching for similar patches in adjacent frames.
VBM4D \cite{maggioni2012video} further generalizes this idea to spatio-temporal volumes.
VNLB \cite{arias2015towards} extends the image-based NLB algorithm \cite{lebrun2013nonlocal} in a similar manner.
The first data-driven method \cite{chen2016deep} exploits recurrent neural networks for temporal information,
but its performance is not satisfactory.
Recently, Davy et al. proposed VNLnet \cite{davy2019vnlnet} which augments DnCNN with a spatio-temporal nearest neighbor module,
allowing non-local patch search and CNN to be combined.
Tassano et al. proposed FastDVDnet \cite{tassano2019fastdvdnet}, which employs a cascade of U-shaped encoder-decoder architectures \cite{ronneberger2015u} and performs implicit motion estimation.
However, these methods are restricted to their training noises
and generalize poorly on other noises.
In \cite{ehret2019joint,mildenhall2018burst}, the related problem of burst denoising is addressed,
but these methods are not tailored for videos containing large motions.

\noindent \textbf{Blind denoising.}
Many efforts have been devoted to blind denoising lately.
The first line of research constructs noisy-clean pairs to train deep architectures. 
Methods were developed for acquiring the GT of real photographs:
many datasets \cite {abdelhamed2018high,anaya2018renoir,plotz2017benchmarking,xu2018real} were proposed,
and deep architectures tailored for realistic noises were trained \cite{anwar2019real,song2019dynamic,yue2019variational}.
However, high quality GT videos are harder to obtain comparing to images, 
preventing these methods from being applied to temporal denoising.
Meanwhile, analytic models were proposed for simulating realistic noises:
CBDNet \cite{guo2019toward} incorporates in-camera processing pipelines into noise modeling;
ViDeNN \cite{claus2019videnn} considers photon shot and read noises.
Nevertheless, these methods are still dependent on their respective analytic models which exhibit inevitable bias,
and may not generalize well to other noises.
Our method does not require expensive video GT datasets, and only imposes weak statistical assumptions on noise attributes.

The second line of research trains self-adaptive denoisers with noisy data alone without clean counterparts.
\textsc{Noise2noise} \cite{lehtinen2018noise2noise} observes that under mild conditions on the noise distribution, 
an image denoiser can be trained with noisy pairs that have the same GT and independent noise realizations.
\textsc{Frame2frame} \cite{ehret2019model} constructs such noisy pairs from videos using optical flow.
Taking a step further, \textsc{noise2self} \cite{batson2019noise2self} and \textsc{noise2void} \cite{krull2019noise2void} 
propose to train image denoisers using single noisy images as both input and target, 
where part of the input is removed from the output's receptive field to avoid learning a degenerate identity function.
However, these methods do not outperform \textsc{noise2noise} as less information is available during training.
Our proposed twin sampler is reminiscent of these methods in the decoupling of inputs from targets;
but instead of removing elements,
we replace elements without changing semantic content.
It tackles noise overfitting due to the temporal redundancy in static scenes, 
a problem that previous image-based methods do not encounter.
\section{Methods}
\subsection{Background}
In discriminative learning paradigm, 
noisy inputs $\bm y_i$ are synthesized from clean images $\bm x_i$ as
$\bm y_i = \bm x_i + \bm n_i$,
where $\bm n_i$ follows some analytic noise distribution, e.g. AWGN.
The \textsc{noise2noise} framework uses noisy-noisy pairs $(\bm y_i, \bm y_i')$ instead,
dispensing with the need for explicit noise modeling or noisy-clean datasets.
Specifically,
it assumes the noisy pair satisfies
\begin{align}
	\bm y_i = \bm x_i + \bm n_i,~\bm y_i' = \bm x_i + \bm n_i',
\end{align}
i.e. they share the same GT.
A neural network $g_\theta$ with weights $\theta$ is then trained by minimizing the empirical risk:
\begin{align} \label{eq:n2n-criterion}
 \argmin_\theta~\mathbb{E}_{\bm y_i, \bm y_i'}[\ell(g_\theta(\bm y_i), \bm y_i')],
\end{align}
where $\ell$ is, say, $L_2$ loss.
With a sufficiently large training set, 
the network $g_\theta$ learns to approximate the optimal estimator $g^*$,
which is $\mathbb{E}[\bm y_i'\mid \bm y_i]$ according to Bayesian decision theory.
If the noise distribution satisfies 
\begin{align} \label{eq:n2n-assumption}
\mathbb{E}[\bm y_i'\mid \bm y_i] = \mathbb{E}[\bm x_i\mid \bm y_i],
\end{align}	
i.e. the noise $\bm n_i'$ preserves mean, the optimal estimator $g^*$ would be the same as if $\bm y_i'$ was replaced by $\bm x_i$ in the training criterion (\ref{eq:n2n-criterion}).
In other words, the network has the same optimal weights $\theta$ as if it was trained using noisy-clean pairs $(\bm y_i, \bm x_i)$.
The same property holds for $L_1$/$L_0$ loss under median/mode-preserving noises.

Such noisy pairs are still difficult to obtain from real data.
\textsc{noise2void} proposes to use $(\bm y_i, \bm y_i)$ to train 
a ``blind-spot'' architecture that removes input pixels from the output's receptive field at same spatial coordinates.
However, it is incompatible with many state-of-the-art methods \cite{davy2019vnlnet,tassano2019fastdvdnet} that directly add inputs to outputs for residual prediction,
and is shown in \cite{krull2019noise2void} to perform worse than \textsc{noise2noise} and traditional methods e.g. BM3D.

Alternatively, \textsc{frame2frame} proposes to use videos to construct such noisy pairs.
They assume that consecutive frames $\{\bm y_{i-1}, \bm y_i\}$ are both observations of the same clean signal $\bm x_i$, except that $\bm y_{i-1}$ is transformed by motion.
Optical flow is computed from $\{\bm y_{i-1}, \bm y_i\}$ and used to warp $\bm y_{i-1}$ to become $\bm y_i'$, 
such that $\bm y_i'$ and $\bm y_i$ are aligned.
Occluded pixels are inferred from the divergence of optical flow and then excluded from the loss (\ref{eq:n2n-criterion}).
Image denoisers trained with noisy pairs from a video can be used to denoise that specific video in a frame-by-frame manner.

\subsection{Optical Flow Refinement} 
\textsc{Frame2Frame} learns image denoisers that take single frames $\bm y_i$ as inputs.
 To incorporate temporal information,
 video denoising networks include the adjacent frames of $\bm y_i$ in their inputs as well, denoted by $Y_i$:
 \begin{align} \label{eq:network-inputs}
 Y_i \coloneqq \{..., \bm y_{i-1}, \bm y_i, \bm y_{i+1}, ...\},
 \end{align}
 where $\coloneqq$ denotes definition/assignment.
The optical flow estimator plays a critical role in improving denoising performance.
There are two major drawbacks of using existing optical flow estimators.
(i) They are usually trained with clean image pairs,
which can perform worse on noisy inputs.
(ii) Most optical flow estimators are designed to match the GT flow, regardless of the image content.
As the goal of optical flow warping is to correctly align adjacent frames, 
we would like to penalize misalignments that result in large pixel differences.
For example, flow error in homogeneous regions (e.g. a whiteboard) 
will not cause violation of the \textsc{noise2noise} assumption, 
and should not be penalized as much as heterogeneous regions.

To solve (i), one might use synthetic noise to train a noise-robust optical flow estimator.
However, in our problem setup such method is not viable as noise models are unknown.
Instead, we perform \textit{online denoising} before estimating optical flow,
using the very denoiser $g_\theta$ we are training.
Intuitively, the denoiser $g_\theta$ and the optical flow quality
evolve with each other:
throughout training, $g_\theta$ produces progressively cleaner frames 
that improve optical flow estimation,
which in turn helps to train $g_\theta$ via better alignment.
Formally, suppose an optical flow estimator $\Gamma$ computes the optical flow from $\bm a$ to $\bm b$ as $\Gamma(\bm a, \bm b)$.
The forward and backward flow $\bm w^f$, $\bm w^b$ between $\bm y_{i-1}$ and $\bm y_i$ are computed as
\begin{align}
\begin{split} \label{eq:optical-flow-compute}
	\bm w^f \coloneqq \Gamma(g_\theta(Y_{i-1}), g_\theta(Y_{i})),~
	\bm w^b \coloneqq \Gamma(g_\theta(Y_{i}), g_\theta(Y_{i-1})).
\end{split}
\end{align}
\begin{figure}[t]
\begin{center}
   \includegraphics[width=0.8\linewidth]{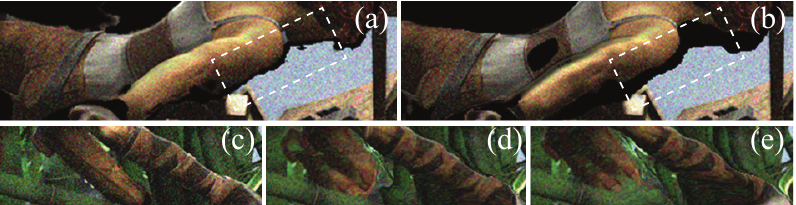}
\end{center}
   \caption{
Comparison between the original loss and (\ref{eq:regularized-optical-loss}) on PWC-net \cite{sun2018pwc}.
(a) and (b): warped images with inferred occlusion (black) using 
(\ref{eq:regularized-optical-loss}) and the original loss.
(a) has smaller inferred occlusion.
(c): reference frame.
(d) and (e): warped frames using 
(\ref{eq:regularized-optical-loss}) and original loss.
(d) matches (c) more faithfully
   }
\label{fig:flow}
\end{figure}
To solve (ii), we use a warping loss to regularize the training of $\Gamma$.
This loss directly penalizes pixel difference after alignment.
Suppose the GT flow from $\bm a$ to $\bm b$ is $\bm w$,
and the original loss function is $\mathcal{L}_\textit{orig}=\mathcal{L}(\Gamma(\bm a, \bm b), \bm w)$,
which only considers flow error.
We use the following hybrid loss instead (a comparison is shown in Fig.~\ref{fig:flow}):
\begin{align} \label{eq:regularized-optical-loss}
	\mathcal{L}_\textit{orig}
	 +\lambda \big\lVert (1 - \bm o_a)\odot \big(\bm a - \text{warp}(\bm b, \Gamma(\bm a, \bm b))\big)\big\rVert_2^2,
\end{align}
where ``warp'' represents the inverse warping function, and $\lambda$ is a hyper-parameter that controls the balance between these two terms.
The GT occlusion map $\bm o_a$ is used to exclude occluded regions where no alignment can be achieved.
As such, training with this loss requires datasets that contain GT occlusion maps, 
e.g. FlyingChairs2 \cite{ISKB18} and Sintel \cite{butler2012sintel}.
Meister et al. \cite{meister2018unflow} used the warping loss
to train $\Gamma$ without GT flow.
In our scenario, we found the GT flow to be a useful guidance,
hence the warping loss is only used as a regularization term in (\ref{eq:regularized-optical-loss}).

\subsection{Correspondence Management} \label{subsec:correspondence-handling}
\begin{figure*}[t]
\begin{center}
   \includegraphics[width=\linewidth]{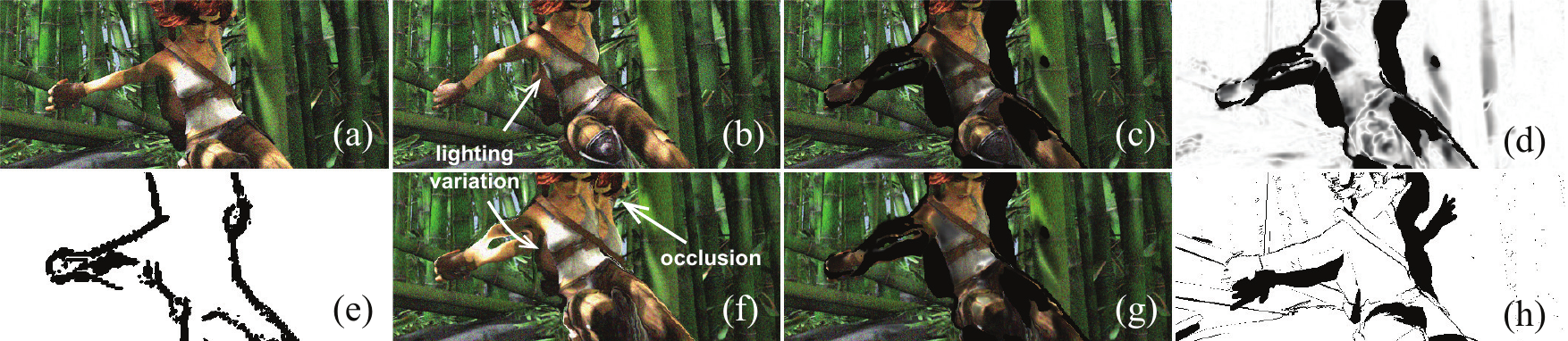}
\end{center}
   \caption{
   An illustration of correspondence management:
   (a) frame $\bm y_{i-1}$; (b) frame $\bm y_i$; 
   (f) $\bm y_i'$ (frame $\bm y_{i-1}$ warped towards $\bm y_i$);
   (d) inferred occlusion mask (solid black) and lighting variation (gray) from Sec. \ref{subsec:correspondence-handling};
   (c) and (g): multiply (b) and (f) with mask (d), respectively;
   (e) inferred occlusion mask based on flow divergence;
   (h) GT occlusion.
   Compare (b) and (f) to observe occlusion and lighting variation.
   Further compare with (c) and (g) to see that the mask (d) effectively covers these outlier pixels
   }
\label{fig:correspondence}
\end{figure*}

Let us further analyze the scenarios where the \textsc{noise2noise} assumption (\ref{eq:n2n-assumption}) fails in the case of multiple frame input $Y_i$.
The assumption requires that 
$\mathbb{E}[\bm y_i'\mid Y_i] = \mathbb{E}[\bm x_i\mid Y_i]$,
where $\bm y_i'$ is obtained by warping $\bm y_{i-1}$ towards $\bm y_i$ using optical flow.
It fails if (i) a pixel in $\bm y_{i-1}$ has no correspondence in $\bm y_i$,
or (ii) the corresponding pixels in $\bm y_{i-1}$ and $\bm y_i$ have different GT values.
Occlusion causes (i),
while lighting variation leads to (ii), see Fig.~\ref{fig:correspondence} (b) and (f).

In \textsc{frame2frame}, occlusion is detected by checking 
if the divergence of optical flow exceeds a threshold.
As it turns out, we can use the forward and backward optical flow 
computed in the previous subsection to derived a better occlusion mask, see Fig.~\ref{fig:correspondence} (d) and (e).
The forward-backward consistency assumption \cite{sundaram2010dense} states that 
the forward flow of a non-occluded pixel and the backward flow of its corresponding pixel in the next frame should be opposite numbers.
Meister et al. \cite{meister2018unflow} used this property to regularize unsupervised training of optical flow.
Here we use this property to directly infer occlusion.
Specifically, let $\bm p$ denote a pixel coordinate in $\bm y_i$;
we can compute a binary map $\bm o_i$ to mark if $\bm p$ is occluded in the previous frame $\bm y_{i-1}$: let $\bm o_i(\bm p) \coloneqq 0$ (not occluded) if 
\begin{align}
\begin{split} \label{eq:def-occlusion-map}
	\big\lVert \bm w^b(\bm p) + \bm w^f\big(\bm p + \bm w^b(\bm p)\big)\big\rVert_2^2 < \alpha_1 \bigg(\big\lVert \bm w^b(\bm p)\big\rVert_2^2 + \big\lVert \bm w^f\big(\bm p + \bm w^b(\bm p)\big)\big\rVert_2^2\bigg)+\alpha_2,
\end{split}
\end{align}
otherwise $\bm o_i(\bm p) \coloneqq 1$ (occluded), where $\alpha_1$,$\alpha_2$ are hyper-parameters specifying relative and absolute thresholds.

While occlusion masks are binary, lighting variation 
is quantitative by nature.
We can use the difference between corresponding pixels, 
e.g. pixels at same coordinates of $\bm x_i$ and $\bm x_i'$ ($\bm x_{i-1}$ warped towards $\bm x_i$),
to quantify lighting variation.
However, individual pixels can have large variance due to noise and randomness.
To improve robustness, we instead compare the average intensity of patches centered at corresponding pixels.
Using a $5\times 5$ box filter $\kappa_5$,
the patch difference can be computed efficiently (e.g. on GPUs) as $|\kappa_5 * (\bm x_i - \bm x_i')|$, where $*$ denotes convolution.
Again, occluded pixels should be excluded from the patch.
This can be done by a point-wise product between $\bm x_i - \bm x_i'$ and the non-occlusion map $1 - \bm o_i$,
followed by proper normalization. 
Formally, the lighting variation $\bm l_i$ of pixels in $\bm x_i$ with respect to corresponding pixels in $\bm x_{i-1}$ is computed as
\begin{align} \label{eq:def-illumination-change}
	\bm l_i \coloneqq \frac{\big|\kappa_5 *[(\hat{\bm x}_i - \hat{\bm x}'_i)\odot (1-\bm o_i)]\big|}
	{\kappa_5 * (1 - \bm o_i) + \epsilon},
\end{align}
where $\hat{\bm x}_i$ and $\hat{\bm x}'_i$ represent our estimates of the GT quantities $\bm x_i$ and $\bm x_i'$, $\odot$ and fraction denote point-wise product and division, and the denominator is a normalization factor that contains a small positive $\epsilon=10^{-6}$ to prevent division by zero.
The evaluation of lighting variation (\ref{eq:def-illumination-change}) can also benefit from online denoising,
so as to prevent the pixel difference due to noise from being misinterpreted as lighting variation.
To do so, we define the clean signal estimates $\hat{\bm x}_i$ and $\hat{\bm x}'_i$ in (\ref{eq:def-illumination-change}) as:
\begin{align}
\begin{split} \label{eq:clean-signal-estimates}
\hat{\bm x}_i \coloneqq g_\theta(Y_i),~
\hat{\bm x}'_i \coloneqq \text{warp}(g_\theta(Y_{i-1}), \bm w^b).
\end{split}
\end{align}
This reduces the amount of noise in clean signal estimates to further improve robustness.

\subsection{Twin Sampler} \label{subsec:twin-sampler}
A naive extension of \textsc{frame2frame} to video denoisers
 is to train with $(Y_i, \bm y_i')$.
 Unfortunately this leads to noise overfitting in static regions.
Since the optical flow is almost zero in these regions ($\bm y_i' \approx \bm y_{i-1}$),
the target $\bm y_i'$ becomes a part of the input $Y_i$.
The network can simply learn to reproduce that part (the previous frame $\bm y_{i-1}$),
leading to noisy prediction.
A visual example is provided later in Fig.~\ref{fig:hyperparams} (right).
To avoid this, one might consider  to use a frame $\bm y_j \notin Y_i$ for computing $\bm y_i'$.
However this is impractical as state-of-the-art video denoising networks can have $Y_i$ that spans a large temporal window (up to 7 in both directions \cite{davy2019vnlnet}),
and frames that are too far from $\bm y_i$ simply cannot be aligned with.



We propose a twin sampler that not only effectively solves this problem, 
but also brings additional benefits.
Our first step is to replace $\bm y_{i-1}$ in $Y_i$ with a warped frame as well.
A toy example that illustrates this idea is given below.
Suppose the network $g_\theta$ originally takes three frames $Y_2=\{\bm y_1, \bm y_2, \bm y_3\}$  as input
to denoise the middle frame $\bm y_2$.
Using estimated optical flow,
we warp $\bm y_3$ to align with $\bm y_2$, yielding $\bm y_{3\rightarrow 2}$;
similarly, $\bm y_2$ is warped to align with $\bm y_3$, yielding $\bm y_{2\rightarrow 3}$.
The new input is $Y_2' = \{\bm y_1, \bm y_2, \bm y_{2\rightarrow 3}\}$ which resembles the original input $\{\bm y_1, \bm y_2, \bm y_3\}$
semantically.
The target is still $\bm y_{3\rightarrow 2}$ which is another noisy observation of $\bm x_2$.
The key is that the input $Y_2'$ and the target $\bm y_{3\rightarrow 2}$ do not share sources:
pixels in $Y_2'$ originate from $\bm y_1$ and $\bm y_2$,
and pixels in $\bm y_{3\rightarrow 2}$ originate from $\bm y_3$.
As such, a degenerate mapping that produces part of the input will not be learned.
Also, since $Y_2'$ keeps the semantic form of the original input $Y_2$,
the network's interpretation remains the same thus no change is required during inference time.

As simple as it may seem, this method comes with two convenient byproducts.
Firstly,
since the forward and backward flow between $\bm y_2$ and $\bm y_3$ have been computed during the above process,
the occlusion mask in Section \ref{subsec:correspondence-handling} can be derived with little additional cost.
Secondly, another noisy pair, $(Y_3'=\{\bm y_{3\rightarrow 2}, \bm y_3, \bm y_4\}, \bm y_{2\rightarrow 3})$, can be immediately constructed without additional optical flow estimation/warping.
The input $Y_3'$ resembles $Y_3=\{\bm y_2, \bm y_3, \bm y_4\}$ semantically, the target $\bm y_{2\rightarrow 3}$ is another noisy observation of $\bm x_3$,
and no sources are shared between them.
This method is dubbed a ``twin sampler'' due to the fact that constructed samples are always grouped in pairs corresponding to consecutive frames.

Formally, let $\bm y_{i\rightarrow j}$ denote the frame obtained by warping $\bm y_i$ towards $\bm y_j$.
Suppose the network input takes the general form (\ref{eq:network-inputs}).
The twin sampler first computes 
\begin{align}
\begin{split} \label{eq:warp}
	\bm y_{(i-1)\rightarrow i} \coloneqq \text{warp}(\bm y_{i-1}, \bm w^b),~
	\bm y_{i\rightarrow (i-1)} \coloneqq \text{warp}(\bm y_i, \bm w^f), 
\end{split}
\end{align}
Then, two noisy pairs are constructed as 
\begin{align}
\Big(Y_{i-1}'\coloneqq Y_{i-1}\setminus\{\bm y_i\}\cup \{\bm y_{(i-1)\rightarrow i}\}, \bm y_{i\rightarrow (i-1)}\Big), \label{eq:twin-sample-1} \\
\Big(Y_i'\coloneqq Y_i\setminus\{\bm y_{i-1}\}\cup \{\bm y_{i\rightarrow (i-1)}\}, \bm y_{(i-1)\rightarrow i}\Big). \label{eq:twin-sample-2}
\end{align}

The occlusion map $\bm o_i$ and lighting variation $\bm l_i$
are used to adjust the loss $\ell$ in the training criterion (\ref{eq:n2n-criterion}).
For the noisy pair (\ref{eq:twin-sample-2}), its associated loss is 
\begin{align} \label{eq:loss-mask}
\begin{split}
	\ell\big(g_\theta(Y_i') \odot \bm \gamma, \bm y_{(i-1)\rightarrow i} \odot \bm \gamma\big)~\text{where}~\bm \gamma = (1 - \bm o_i) \odot \xi(\bm l_i),
\end{split}
\end{align}
and $\xi$ is a non-linear function with a hyper-parameter $\alpha_3$ that maps its input to range $(0,1]$:
\begin{align} \label{eq:def-xi}
	\xi(l) \coloneqq \exp(-\alpha_3 l).
\end{align}
Intuitively, occluded pixels do not contribute to the loss,
and pixels with drastic lighting variation contribute less to the loss.
Therefore, our loss function guides the network to learn from pixels that are properly aligned and satisfy the assumption (\ref{eq:n2n-assumption}).
For the other noisy pair (\ref{eq:twin-sample-1}),
its occlusion map $\bm o_{i-1}$, lighting variation $\bm l_{i-1}$ and associated loss are computed similarly,
with $\bm w^f/\bm w^b$ exchanged in (\ref{eq:def-occlusion-map}) and $i/i-1$ exchanged in (\ref{eq:def-illumination-change})(\ref{eq:loss-mask}).

\begin{algorithm}[t]
\caption{Our training procedure for each mini-batch. The optical flow network $\Gamma$ has been trained with loss (\ref{eq:regularized-optical-loss}).}
\label{code:algorithm}
\begin{algorithmic}[1]
\While{the batch is not full; select a random $i$ and}
\State Construct original inputs $Y_{i-1}$ and $Y_i$ by (\ref{eq:network-inputs}).
\State Compute optical flow $\bm w^f, \bm w^b$ and clean signal estimates $\hat{\bm x}_i$ and $\hat{\bm x}'_i$ by (\ref{eq:optical-flow-compute})(\ref{eq:clean-signal-estimates}).
\State Compute backward occlusion map $\bm o_i$ and lighting variation $\bm l_i$ by (\ref{eq:def-occlusion-map})(\ref{eq:def-illumination-change}). Forward $\bm o_{i-1}$ and $\bm l_{i-1}$ are computed similarly with $i-1$ and $i$ exchanged.
\State Construct the final input and target from (\ref{eq:warp})(\ref{eq:twin-sample-1})(\ref{eq:twin-sample-2}), crop the above quantities at same spatial locations and add them to the mini-batch. \textbf{end while}\label{code:crop}
\EndWhile
\State Compute loss (\ref{eq:loss-mask}) and update weights $\theta$ with backprop.\label{code:backprop}
\end{algorithmic}
\end{algorithm}

\subsection{Summary} \label{subsec:dae}
The pseudocode summarizing the above procedures is shown in Algorithm \ref{code:algorithm}.
We train the network $g_\theta$ using mini-batches,
each consists of multiple noisy pairs.
Since the noisy video can have very high resolutions,
in line \ref{code:crop} we crop these pairs as well as their associated occlusion maps and lighting variations to a fixed size.
This allows us to use large batch sizes regardless of the video's resolution.
All related computations can efficiently run on GPUs.

Following \cite{ehret2019model}, $\theta$ is initialized by pretraining with synthetic AWGN on clean datasets.
Since GT is utilized in this pretraining but not in our method,
if the actual noise is very close to AWGN,  
the initial model operates in an ideal test situation and is likely to perform very well.
Our final trick is to use a denoising autoencoder (DAE) to detect if this happens.
According to Alain et al. \cite{alain2014regularized},
the reconstruction error of a DAE $r(\bm y)$ trained with infinitesimal AWGN
satisfies
$r(\bm y) - \bm y \approx \sigma^2\nabla_{\bm y}\log \Pr (\bm y)$,
where $\Pr(\bm y)$ is the data distribution of GT.
If $\bm y$ contains little noise, 
it is close to the GT manifold, which implies $\Pr(\bm y)$ is 
close to its local maximum
and the reconstruction error will be small.
Therefore, we can use $\lVert r(\bm y)-\bm y\rVert$ as a rough indicator of the cleanliness of $\bm y$.
Using a DAE trained on ImageNet \cite{ILSVRC15},
we compute the reconstruction error of denoised video frames
before and after \textsc{noise2noise} training.
If the average error magnitude does not decrease significantly (50\%),
the initial model will be kept. 

\section{Experimental Results}
\subsection{Data and Implementation Details}
\noindent \textbf{Data preprocessing.} Due to the lack of reliable methods for obtaining GT of noisy videos,
we use synthetic noises for quantitative experiments as in \cite{ehret2019model},
and demonstrate real noise reduction visually. 
Five distinct synthetic noises are used for testing:
AWGN20 (AWGN with standard deviation $\sigma$=20),
MG (multiplicative Gaussian, where each pixel's value is multiplied by a $\mathcal{N}(1, 0.3^2)$ Gaussian),
CG (correlated Gaussian, where AWGN with $\sigma$=25 is blurred with a 3$\times$3 box filter),
IR (impulse random, where each pixel has 10\% chance to be replaced by a uniform random variable in $[0,255]$),
and JPEG (JPEG compressed Gaussian, where each frame is compressed with 60\% JPEG quality after adding AWGN with $\sigma$=25). 
To mimic realistic scenarios, all pixel values are clipped to range $[0, 255]$ and rounded to nearest integers.

We collect clean videos from three datasets:
Sintel \cite{butler2012sintel}, DAVIS \cite{Perazzi2016} and Derf's collection \cite{derfcollection}. 
The ``clean'' pass of Sintel training set (23 sequences)
are split into 11:4:8, which are used for optical flow training (\texttt{sintel-tr}),
hyper-parameter tuning (\texttt{sintel-val}) and denoising performance evaluation (\texttt{sintel-8}), respectively.
All 30 sequences from the ``test-dev'' split of DAVIS (\texttt{davis-30}) and 7 selected sequences \cite{davy2019vnlnet} from Derf's collection (\texttt{derf-7}) 
are also used for performance evaluation.

\begin{figure}[t]
\begin{center}
   \includegraphics[width=0.6\linewidth]{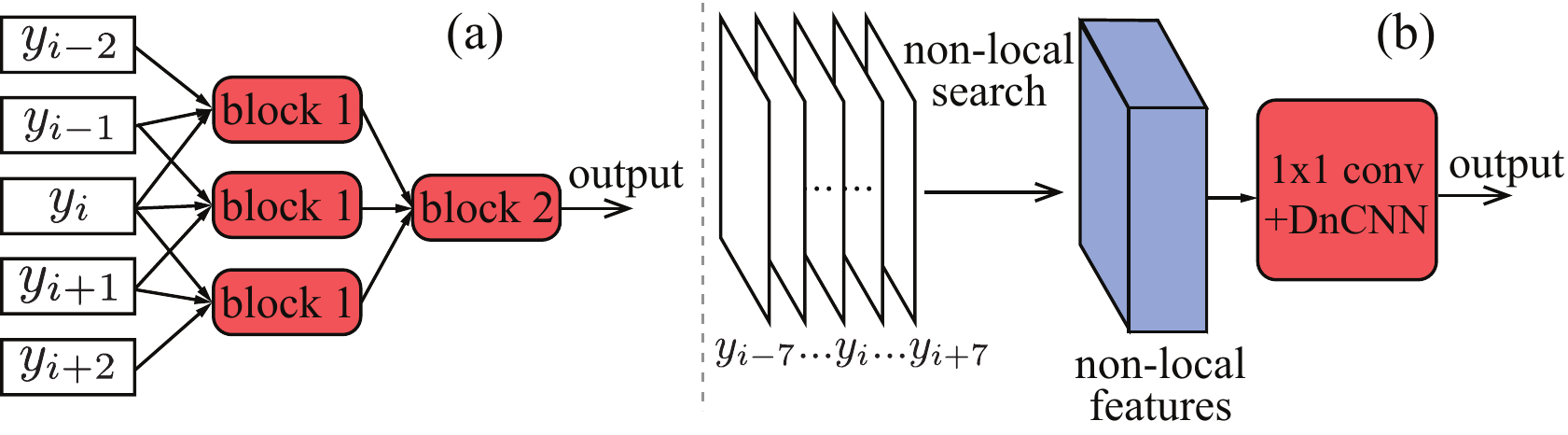}
\end{center}
\vspace*{-10pt}
   \caption{
(a) FastDVDnet takes 5 frames as input and performs two-stage denoising.
(b) VNLnet takes 15 frames as input, which are converted to features using a non-local search module
   }
\label{fig:architecture}
\end{figure}

\noindent \textbf{Implementation.} To demonstrate the generality of our framework,
we apply it to latest video denoising networks with distinct architectures (VNLnet \cite{davy2019vnlnet} and FastDVDnet \cite{tassano2019fastdvdnet}), see Fig.~\ref{fig:architecture}.
The weight used to initialize VNLnet 
is the publicly released version trained on color sequences with AWGN.
The authors of FastDVDnet included noise strength in network input for non-blind denoising.
We train a blind version by removing the noise strength input and repeating the same training procedure.
The noise strength used for training is $\sigma$=20 for both,
which allows the test noise AWGN20 to cover the case where the initial model matches the test noise as discussed in Sec. \ref{subsec:dae}.
\begin{figure}[t]
\begin{center}
   \includegraphics[width=\linewidth]{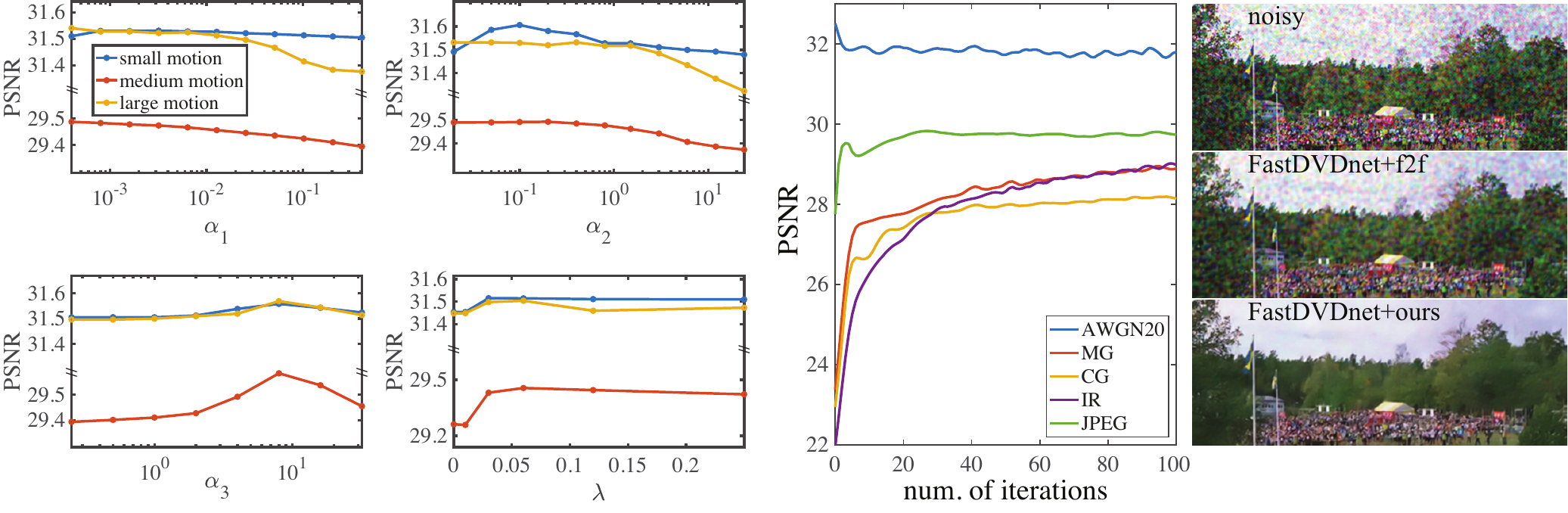}\
\end{center}
\vspace*{-10pt}
\caption{
Left: occlusion masking ($\alpha_1, \alpha_2 < \infty$) is essential for sequences with large motions.
A moderate penalty on lighting variation ($0 < \alpha_3 < \infty$) is optimal.
Regularized optical flow ($\lambda>0$) boosts denoising performance.
Overall the hyper-parameters are not sensitive to small variations.
Middle: denoising a sequence in \texttt{sintel-8}. The performance of FastDVDnet+ours converges steadily as training continues.
Right: \texttt{derf-7} with CG noise. Our method avoids the noise overfitting problem suffered by the naive extension of \cite{ehret2019model}
}
\label{fig:hyperparams}
\end{figure}

We perform random search to determine the best hyper-parameters.
A ``validation noise'' (AWGN with $\sigma$=30) is used
to prevent previous ``test noises'' from being seen. 
The combination that achieves the best average PSNR 
on \texttt{sintel-val} is:
$\alpha_1$=0.0064, $\alpha_2$=1.4 in (\ref{eq:def-occlusion-map});
$\alpha_3$=5.0 in (\ref{eq:def-xi});
 and
$\lambda$=0.06 in (\ref{eq:regularized-optical-loss}).
Within \texttt{sintel-val}, 3 sequences with different motion scales are selected; individual hyper-parameters are varied to study their sensitivities, see Fig.~\ref{fig:hyperparams}.
The loss function $\ell$ in (\ref{eq:n2n-criterion}) is the L1 loss,
which can cope with a wide range of noises according to \cite{ehret2019model}.
We use the Adam optimizer to update weights (Algorithm \ref{code:algorithm} line \ref{code:backprop});
the learning rate is $5\times 10^{-5}$ for FastDVDnet and $2\times 10^{-4}$ for VNLnet;
a fixed batch size 32 is set for both,
and the iteration stops after 100 mini-batches.
The fixed crop size in Algorithm \ref{code:algorithm} line \ref{code:crop} is 96 by 96.
To compute optical flow,
$\Gamma$ is selected as the recently proposed PWC-net \cite{sun2018pwc},
which outperforms many traditional methods and is also faster.
We fine-tune the publicly released model (pretrained with FlyingThings3D \cite{MIFDB16}) using FlyingChairs2 \cite{ISKB18}, ChairsSDHom \cite{IMKDB17} and \texttt{sintel-tr} with loss (\ref{eq:regularized-optical-loss}).
\subsection{Main Results}
Regarding overall performance, we primarily compare with \textsc{frame2frame},
which uses the image denoiser DnCNN as their backbone.
The naive extension of \textsc{frame2frame} to video denoising networks, as described at the beginning of Section \ref{subsec:twin-sampler}, 
also serves as a baseline.
Traditional methods such as VBM4D and VNLB,
as well as some recent blind denoising methods including CBDnet and ViDeNN are also compared.
Note that these recent methods are still trained with noisy-clean pairs,
whose performances are bounded by their training data and noise model assumptions.
For \textsc{frame2frame}, the backbone DnCNN is also initialized by pretraining with AWGN $\sigma$=20.
Since our task is model-blind denoising,
using specialized pretrained model for each test noise is not allowed.
Therefore, for methods that require pretrained weights,
the same publicly released model will be used for all noises.
\begin{table*}
\caption{Average PSNR/SSIM on \texttt{derf-7} and \texttt{davis-30}. 
DnCNN+f2f is the original implementation of \cite{ehret2019model}.
FastDVDnet+f2f and VNLnet+f2f  are naive extensions of \cite{ehret2019model} to video denoisers as described in Sec. \ref{subsec:twin-sampler}.
``X initial'' is the initial model of X pretrained with AWGN,
``X+ours'' is our proposed framework applied to X.
For image denoisers, frame-by-frame denoising is performed
} \label{table:derf}
\begin{center}
\resizebox{\textwidth}{!}{
\begin{tabular}{|l|c|c|c|c|c|c|c|c|c|c|}
\hline
dataset & \multicolumn{5}{c|}{\texttt{derf-7}} & \multicolumn{5}{c|}{\texttt{davis-30}} \\ \hline
noise & AWGN20 & MG & CG & IR & JPEG & AWGN20 & MG & CG & IR & JPEG\\
\hline \hline
VBM4D \cite{maggioni2012video} & 33.23/.896  & \underline{30.11}/\textbf{.850} & 22.70/.471 & 27.55/.743 & 29.44/.793 & 32.79/.890 & 27.28/\textbf{.801} & 22.44/.439 & 26.92/.719 & 28.77/.767\\
VNLB \cite{arias2015towards} & \underline{34.71}/\underline{.916} & 22.16/.589 & 23.20/.509 & 21.56/.498 & \underline{30.55}/\underline{.852} & 34.00/.911 & 19.44/.474 & 23.32/.509 & 21.61/.514 & 30.28/\underline{.853} \\
VDNet \cite{yue2019variational} & 33.05/.893  & 20.22/.463 & 22.04/.452 & 19.64/.387 & 23.73/.486 & 33.58/.912 & 18.13/.380 & 22.11/.446 & 19.90/.399 & 23.92/.479\\
CBDNet \cite{guo2019toward} & 31.91/.866  & 24.12/.646 & 24.19/.582 & 24.76/.613 & 27.84/.717 & 32.45/.890 & 21.95/.564 & 24.62/.598 & 26.30/.682 & 28.56/.754 \\
KPN \cite{mildenhall2018burst} & 20.83/.676 & 17.88/.654 & 18.75/.396 & 19.75/.567 & 21.88/.733 & 21.20/.758 & 20.81/.636 & 18.99/.491 & 21.39/.568 & 21.64/.812 \\
ViDeNN \cite{claus2019videnn} & 33.51/.903 & 20.08/.460 & 22.53/.476 & 18.12/.329 & 23.90/.492 & 34.37/\underline{.924} & 17.98/.379 & 22.54/.471 & 18.15/.320 & 23.99/.477 \\ 
TOFlow \cite{xue2019video} & 32.89/.884 & 23.92/.652 & 23.49/.646 & 27.65/.786 & 24.85/.740 & 31.02/.854 & 23.16/.558 & 23.49/.632 & 26.59/.703 & 24.73/.730  \\
DnCNN+f2f & 31.97/.874 & 28.82/.815 & 27.24/.735 & 29.68/.830 & 29.90/.826 & 31.38/.870 & 26.72/.757 & 27.26/.747 & 28.70/.795 & 29.64/.828 \\
\hline
VNLnet initial & \textbf{34.89}/\textbf{.928} & 22.00/.579 & 23.93/.573 & 20.89/.446 & 28.28/.733 & \textbf{34.67}/\textbf{.927} & 19.28/.465 & 24.07/.571 & 21.07/.467 & 28.32/.721\\
VNLnet+f2f & 28.41/.743 & 26.73/.715 & 25.40/.624 & 29.90/.818 & 27.52/.709 & 28.87/.792 & 25.99/.714 & 25.81/.669 & 29.02/.801 & 27.96/.765\\
VNLnet+ours & \textbf{34.89}/\textbf{.928} & \textbf{30.24}/\underline{.849} & \textbf{30.40}/\textbf{.844} & \textbf{31.34}/\textbf{.838} & \textbf{31.13}/\textbf{.867} & \textbf{34.67}/\textbf{.927} & \textbf{27.81}/.780 & \textbf{29.57}/\textbf{.822} & \textbf{30.84}/\textbf{.860} & \textbf{30.48}/\textbf{.854}\\
\hline
FastDVDnet initial & 33.28/.904 & 22.44/.561 & 23.03/.504 & 21.83/.485 & 27.95/.705 & \underline{34.39}/\textbf{.927} & 19.81/.445 & 23.23/.508 & 22.03/.507 & 28.44/.710\\
FastDVDnet+f2f & 30.55/.839 & 28.23/.779 & 26.08/.673 & 29.54/.817 & 29.06/.792 & 30.07/.847 & 26.78/.772 & 26.63/.726 & 28.72/.805 & 28.81/.810\\
FastDVDnet+ours & 33.28/.904 & 29.62/.834 & \underline{28.90}/\underline{.802} & \underline{30.23}/\underline{.834} & 30.54/.846 & \underline{34.39}/\textbf{.927} & \underline{27.68}/\underline{.795} & \underline{28.60}/\underline{.793} & \underline{29.63}/\underline{.822} & \underline{30.34}/.852 \\
\hline
\end{tabular}
}
\end{center}
\end{table*}

Table~\ref{table:derf} show the overall results on 
\texttt{derf-7} and \texttt{davis-30}.
More details are given in the table caption.
The following observations are clear from the results.
(1) The naive extension of \textsc{frame2frame} to video denoising networks performs even worse than its DnCNN-based version (compare rows with suffix ``+f2f'').
This is due to the noise overfitting problem as discussed before, see Fig.~\ref{fig:hyperparams} (right).
(2) Our method consistently outperforms DnCNN-based \textsc{frame2frame} on both architectures,
achieving 0.6-3.2dB PSNR gain (``DnCNN+f2f'' v.s. rows with suffix ``+ours'').
This proves that our method successfully leverages the capability of video denoising networks to utilize temporal information.
(3) Comparing to other existing methods, our method achieves state-of-the-art results on 
removing model-blind noises.
This is expected as those other methods are designed for their respective noise models, and do not have the capability to adapt to existing video noise if their models are violated.
(4) According to columns ``AWGN20'',
our method can effectively detect 
if the actual noise is close to the training noise used to initialize weights 
and select the appropriate model.

\begin{figure}[t]
\begin{center}
   \includegraphics[width=\linewidth]{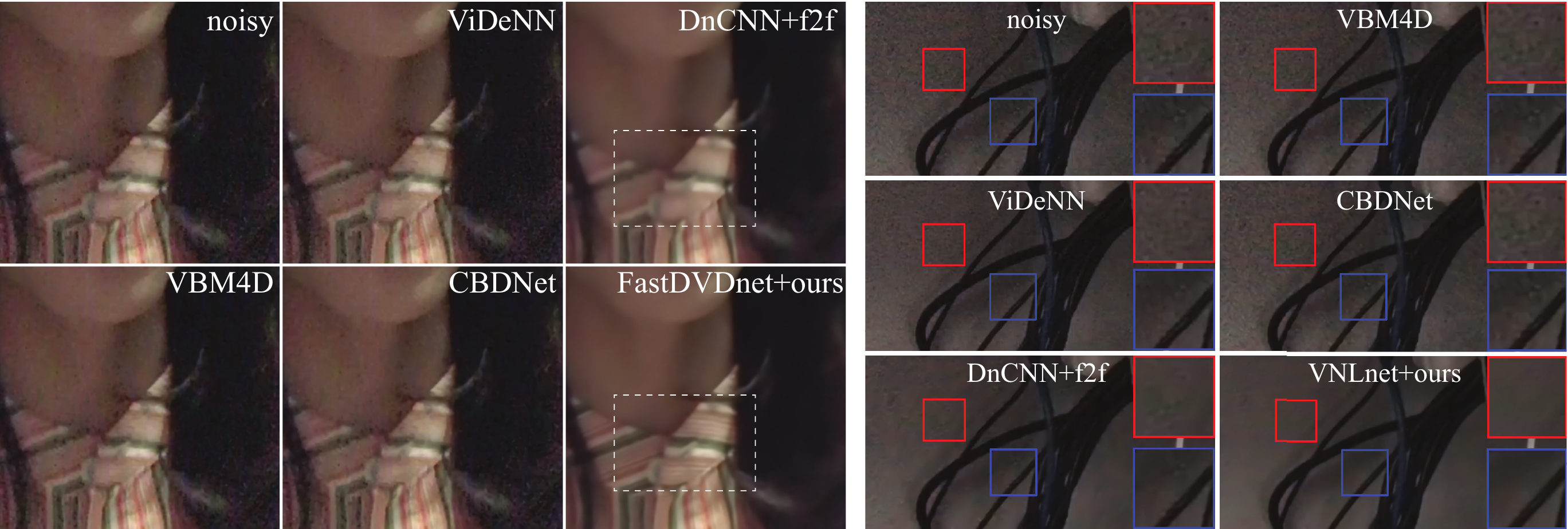}
\end{center}
\vspace*{-10pt}
   \caption{
Denoising real videos captured with a front-facing camera (left) or low lighting condition (right).
Model-based methods, even though targeting realistic noises (mid column), fail in this case.
Our method outperforms image-based \textsc{frame2frame} by incorporating temporal information
   }
   \vspace*{-5pt}
\label{fig:subjective2}
\end{figure}

We also use mobile devices to capture several real sequences,
whose subjective denoising results are shown in Fig.~\ref{fig:subjective2}.
Models trained with synthetic data, even though equipped with realistic noise models, 
fail to remove these real noises,
revealing the limitation of model-based approaches.
Our method produces clean results and outperforms \cite{ehret2019model} on both architectures.

\subsection{Ablation Studies}

Table \ref{table:ablation} shows the detailed breakdown of our method's performance on dataset \texttt{sintel-8}.
Due to space constraints, the complete breakdown for other noises are given in the supplementary.
Since \texttt{sintel-8} contains GT optical flow and occlusion,
we can compare with two oracles that exploit these GT:
row 8 employs \textsc{frame2frame},
while row 9 employs our twin sampler.
\begin{table*}
\caption{Average PSNR/SSIM on \texttt{sintel-8}.
``ts'': twin sampler. If twin sampler is disabled, the naive extension of \textsc{frame2frame} is used.
``occ'': occlusion inference method. ``div''/``ofc'': occlusion is inferred based on optical flow divergence/consistency.
``lv'':  lighting variation. If it is disabled, lighting variations $\bm l_{i-1}, \bm l_i$ are set to 0.
``od'': online denoising. 
``wl'': warping loss regularizer. If it is disabled, $\Gamma$ is fine-tuned on the same datasets, but with the original loss.
``dae'': DAE module as described in Sec. \ref{subsec:dae}} \label{table:ablation}
\begin{center}
\resizebox{\textwidth}{!}{
\begin{tabular}{|c|c|c|c|c|c|c|c|c|c|c|c|c|c|c|}
\hline
& \multicolumn{6}{c|}{components} & \multicolumn{4}{c|}{VNLnet} & \multicolumn{4}{c|}{FastDVDnet}  \\ \cline{2-15}
 & ts & occ & lv & od & wl & dae & AWGN20 & AWGN40 & MG & JPEG & AWGN20 & AWGN40 & MG & JPEG \\ \cline{2-15}
\hline \hline
1 & \xmark & div & \xmark & \xmark & \xmark & \xmark & 29.06/.763 & 27.00/.662 & 27.96/.759 & 28.30/.748 & 30.60/.848 & 28.28/.768 & 29.02/.825 & 30.00/.814 \\
2 & \cmark & div & \xmark & \xmark & \xmark & \xmark & 32.76/.886 & 29.62/.802 & 30.50/.842 & 31.14/.850 & 32.81/.890 & 29.55/.809 & 30.45/.854 & 31.11/.853 \\
3 & \cmark & ofc & \xmark & \xmark & \xmark & \xmark & 33.49/.899 & 30.10/.814 & 30.95/.849 & 31.61/.860 & 33.26/.897 & 29.96/.817 & 30.98/.864 & 31.48/.859 \\
4 & \cmark & ofc & \cmark & \xmark & \xmark & \xmark & 33.68/.902 & 30.20/.818 & 31.07/.851 & 31.74/.863 & 33.39/.898 & 30.06/.820 & 31.03/.865 & 31.54/.861 \\
5 & \cmark & ofc & \cmark & \cmark & \xmark & \xmark & 33.78/.902 & 30.40/.825 & 31.34/.864 & 31.83/.864 & 33.59/.903 & 30.17/.823 & 31.09/.866 & 31.61/.863 \\
6 & \cmark & ofc & \cmark & \cmark & \cmark & \xmark & 34.04/.906 & 30.52/.828 & 31.43/.865 & 31.95/.866 & 33.81/.906 & 30.26/.824 & 31.13/.866 & 31.67/.864 \\
7 & \cmark & ofc & \cmark & \cmark & \cmark & \cmark & 35.45/.925 & 30.52/.828 & 31.43/.865 & 31.99/.867 & 34.92/.922 & 30.26/.824 & 31.13/.866 & 31.70/.864 \\ \hline
8 & \xmark & GT & \xmark & \multicolumn{2}{c|}{GT flow} & \xmark & 30.62/.768 & 27.02/.623  & 29.04/.767 & 29.12/.731 & 32.33/.847 & 28.53/.727 & 29.55/.798 & 30.12/.783 \\
9 & \cmark & GT & \xmark & \multicolumn{2}{c|}{GT flow} & \xmark & 34.28/.918 & 30.54/.830 & 31.37/.862 & 31.95/.873 & 34.14/.916 & 30.73/.841 & 31.50/.879 & 31.92/.875 \\ \hline
10 & \multicolumn{6}{c|}{initial model} & 35.45/.925 & 23.94/.450 & 23.68/.647 & 29.50/.741 & 34.92/.922 & 23.94/.436 & 23.87/.611 & 29.15/.703 \\
\hline
\end{tabular}
}
\end{center}
\vspace*{-5pt}
\end{table*}

The \textbf{twin sampler} offers the most significant contribution,
as PSNR is improved by 1.1-3.7dB (row 1 v.s. 2).
It also benefits the oracle by 1.8-3.7dB (row 8 v.s. 9).
The gap between row 2 and 9 (0.8-1.5dB) results from inaccurate correspondence.
For occlusion masking, \textbf{flow consistency} clearly outperforms flow divergence,
achieving 0.4-0.7dB PSNR gain (row 2 v.s. 3).
By considering \textbf{lighting variation},
PSNR is improved by 0.06-0.2dB (row 3 v.s. 4).
By improving optical flow quality, \textbf{online denoising} and \textbf{warping loss}
contribute roughly equally, providing 0.1-0.4dB PSNR gain in total (row 4/5/6).
The \textbf{DAE} module (row 7) selects the better model between row 6 and 10. 
When the test noise matches pretraining noise (AWGN20), the initial model is indeed selected.
Even the oracle (row 9) does not match the initial model,
as these state-of-the-art video denoisers are very powerful in ideal test situations once trained.
More notably, for JPEG noise the PSNR of row 7 is higher than both row 6 and 10. 
This is due to the JPEG noise being moderately close to AWGN20 (see Fig.~\ref{fig:hyperparams} middle),
and the better model can be either row 6 or 10, depending on the video content.
Overall, by combining all components, the gap between row 2 and 9 is significantly reduced on FastDVDnet and even surpassed on VNLnet (due to lighting variation).

To demonstrate our method's \textbf{robustness to noise levels}, 
Table \ref{table:ablation} also lists the detailed breakdown on two different Gaussian noises: AWGN20 ($\sigma$=20) and AWGN40 ($\sigma$=40).
It can be seen that our proposed method achieves consistent improvement across different noise strengths.

\section{Conclusion}
We have presented a general framework for adapting video denoising networks to model-blind noises without utilizing clean signals.
The twin sampler not only resolves the overfitting problem suffered by the naive extension of image-based methods,
but also operates efficiently by reusing estimated optical flow.
The rest components further boost denoising performance via occlusion masking, lighting variation penalty and optical flow refinement.
Our results indicate that in order to train a video denoiser with only noisy data,
one shall look at frame differences and similarities simultaneously:
noise attributes can be learned from the former, 
while temporal information can be extracted from the latter. 
Our method consistently outperforms the prior art by 0.6-3.2dB PSNR on multiple noises and datasets.
The significance of our method is also reflected in its generality,
as it is successfully applied to multiple latest architectures.

\clearpage
%
%
\bibliographystyle{splncs04}
\bibliography{main}

\begin{thebibliography}{10}
\providecommand{\url}[1]{\texttt{#1}}
\providecommand{\urlprefix}{URL }
\providecommand{\doi}[1]{https://doi.org/#1}

\bibitem{abdelhamed2018high}
Abdelhamed, A., Lin, S., Brown, M.S.: A high-quality denoising dataset for
  smartphone cameras. In: CVPR. pp. 1692--1700 (2018)

\bibitem{alain2014regularized}
Alain, G., Bengio, Y.: What regularized auto-encoders learn from the
  data-generating distribution. JMLR  \textbf{15}(1),  3563--3593 (2014)

\bibitem{anaya2018renoir}
Anaya, J., Barbu, A.: {RENOIR} -- a dataset for real low-light image noise
  reduction. Journal of Visual Communication and Image Representation
  \textbf{51},  144--154 (2018)

\bibitem{anwar2019real}
Anwar, S., Barnes, N.: Real image denoising with feature attention. In: ICCV
  (2019)

\bibitem{arias2015towards}
Arias, P., Morel, J.M.: Towards a {B}ayesian video denoising method. In:
  International Conference on Advanced Concepts for Intelligent Vision Systems.
  pp. 107--117 (2015)

\bibitem{batson2019noise2self}
Batson, J., Royer, L.: Noise2self: Blind denoising by self-supervision. In:
  ICML. pp. 524--533 (2019)

\bibitem{burger2012image}
Burger, H.C., Schuler, C.J., Harmeling, S.: Image denoising: Can plain neural
  networks compete with {BM3D}? In: CVPR. pp. 2392--2399. IEEE (2012)

\bibitem{butler2012sintel}
Butler, D.J., Wulff, J., Stanley, G.B., Black, M.J.: A naturalistic open source
  movie for optical flow evaluation. In: ECCV. pp. 611--625. Springer (Oct
  2012)

\bibitem{chen2018learning}
Chen, C., Chen, Q., Xu, J., Koltun, V.: Learning to see in the dark. In: CVPR.
  pp. 3291--3300 (2018)

\bibitem{chen2016deep}
Chen, X., Song, L., Yang, X.: Deep {RNN}s for video denoising. In: Applications
  of Digital Image Processing (2016)

\bibitem{chen2016trainable}
Chen, Y., Pock, T.: Trainable nonlinear reaction diffusion: A flexible
  framework for fast and effective image restoration. IEEE TPAMI  (2016)

\bibitem{claus2019videnn}
Claus, M., van Gemert, J.: {ViDeNN}: Deep blind video denoising. In: CVPR
  Workshops (2019)

\bibitem{dabov2007video}
Dabov, K., Foi, A., Egiazarian, K.: Video denoising by sparse {3D}
  transform-domain collaborative filtering. In: European Signal Processing
  Conference. pp. 145--149 (2007)

\bibitem{dabov2007image}
Dabov, K., Foi, A., Katkovnik, V., Egiazarian, K.: Image denoising by sparse
  3-{D} transform-domain collaborative filtering. IEEE TIP  (2007)

\bibitem{davy2019vnlnet}
Davy, A., Ehret, T., Morel, J.M., Arias, P., Facciolo, G.: A non-local {CNN}
  for video denoising. In: ICIP. pp. 2409--2413 (2019)

\bibitem{dong2012nonlocally}
Dong, W., Zhang, L., Shi, G., Li, X.: Nonlocally centralized sparse
  representation for image restoration. IEEE TIP  (2012)

\bibitem{ehret2019joint}
Ehret, T., Davy, A., Arias, P., Facciolo, G.: Joint demosaicing and denoising
  by overfitting of bursts of raw images. ICCV  (2019)

\bibitem{ehret2019model}
Ehret, T., Davy, A., Morel, J.M., Facciolo, G., Arias, P.: Model-blind video
  denoising via frame-to-frame training. In: CVPR. pp. 11369--11378 (2019)

\bibitem{gu2014weighted}
Gu, S., Zhang, L., Zuo, W., Feng, X.: Weighted nuclear norm minimization with
  application to image denoising. In: CVPR. pp. 2862--2869 (2014)

\bibitem{guo2019toward}
Guo, S., Yan, Z., Zhang, K., Zuo, W., Zhang, L.: Toward convolutional blind
  denoising of real photographs. In: CVPR. pp. 1712--1722 (2019)

\bibitem{IMKDB17}
Ilg, E., Mayer, N., Saikia, T., Keuper, M., Dosovitskiy, A., Brox, T.:
  Flow{N}et 2.0: Evolution of optical flow estimation with deep networks. In:
  CVPR (2017)

\bibitem{ISKB18}
Ilg, E., Saikia, T., Keuper, M., Brox, T.: Occlusions, motion and depth
  boundaries with a generic network for disparity, optical flow or scene flow
  estimation. In: ECCV (2018)

\bibitem{krull2019noise2void}
Krull, A., Buchholz, T.O., Jug, F.: Noise2void -- learning denoising from
  single noisy images. In: CVPR. pp. 2129--2137 (2019)

\bibitem{lebrun2013nonlocal}
Lebrun, M., Buades, A., Morel, J.M.: A nonlocal {B}ayesian image denoising
  algorithm. SIAM Journal on Imaging Sciences pp. 1665--1688 (2013)

\bibitem{lehtinen2018noise2noise}
Lehtinen, J., Munkberg, J., Hasselgren, J., Laine, S., Karras, T., Aittala, M.,
  Aila, T.: Noise2noise: Learning image restoration without clean data. In:
  ICML. pp. 2971--2980 (2018)

\bibitem{liu2019enhance}
Liu, D., Cheng, B., Wang, Z., Zhang, H., Huang, T.S.: Enhance visual
  recognition under adverse conditions via deep networks. IEEE TIP  (2019)

\bibitem{liu2018image}
Liu, D., Wen, B., Liu, X., Wang, Z., Huang, T.S.: When image denoising meets
  high-level vision tasks: a deep learning approach. In: IJCAI. pp. 842--848
  (2018)

\bibitem{maggioni2012video}
Maggioni, M., Boracchi, G., Foi, A., Egiazarian, K.: Video denoising,
  deblocking, and enhancement through separable 4-{D} nonlocal spatiotemporal
  transforms. IEEE TIP  \textbf{21}(9),  3952--3966 (2012)

\bibitem{mairal2009non}
Mairal, J., Bach, F.R., Ponce, J., Sapiro, G., Zisserman, A.: Non-local sparse
  models for image restoration. In: ICCV. pp. 54--62 (2009)

\bibitem{mao2016image}
Mao, X., Shen, C., Yang, Y.B.: Image restoration using very deep convolutional
  encoder-decoder networks with symmetric skip connections. In: NeurIPS. pp.
  2802--2810 (2016)

\bibitem{MIFDB16}
Mayer, N., Ilg, E., Hausser, P., Fischer, P., Cremers, D., Dosovitskiy, A.,
  Brox, T.: A large dataset to train convolutional networks for disparity,
  optical flow, and scene flow estimation. In: CVPR (2016)

\bibitem{meister2018unflow}
Meister, S., Hur, J., Roth, S.: Un{F}low: Unsupervised learning of optical flow
  with a bidirectional census loss. In: AAAI (2018)

\bibitem{mildenhall2018burst}
Mildenhall, B., Barron, J.T., Chen, J., Sharlet, D., Ng, R., Carroll, R.: Burst
  denoising with kernel prediction networks. In: CVPR. pp. 2502--2510 (2018)

\bibitem{moulin1999analysis}
Moulin, P., Liu, J.: Analysis of multiresolution image denoising schemes using
  generalized {G}aussian and complexity priors. IEEE Transactions on
  Information Theory pp. 909--919 (1999)

\bibitem{Perazzi2016}
Perazzi, F., Pont-Tuset, J., McWilliams, B., {Van Gool}, L., Gross, M.,
  Sorkine-Hornung, A.: A benchmark dataset and evaluation methodology for video
  object segmentation. In: CVPR (2016)

\bibitem{plotz2017benchmarking}
Pl\"{o}tz, T., Roth, S.: Benchmarking denoising algorithms with real
  photographs. In: CVPR. pp. 1586--1595 (2017)

\bibitem{portilla2003image}
Portilla, J., Strela, V., Wainwright, M.J., Simoncelli, E.P.: Image denoising
  using scale mixtures of {G}aussians in the wavelet domain. IEEE TIP  (2003)

\bibitem{ronneberger2015u}
Ronneberger, O., Fischer, P., Brox, T.: U-net: Convolutional networks for
  biomedical image segmentation. In: International Conference on Medical Image
  Computing and Computer-assisted Intervention. pp. 234--241 (2015)

\bibitem{roth2005fields}
Roth, S., Black, M.J.: Fields of experts: A framework for learning image
  priors. In: CVPR (2005)

\bibitem{rudin1992nonlinear}
Rudin, L.I., Osher, S., Fatemi, E.: Nonlinear total variation based noise
  removal algorithms. Physica D: nonlinear phenomena pp. 259--268 (1992)

\bibitem{ILSVRC15}
Russakovsky, O., Deng, J., Su, H., Krause, J., Satheesh, S., Ma, S., Huang, Z.,
  Karpathy, A., Khosla, A., Bernstein, M., Berg, A.C., Fei-Fei, L.: {ImageNet
  Large Scale Visual Recognition Challenge}. IJCV  \textbf{115}(3),  211--252
  (2015). \doi{10.1007/s11263-015-0816-y}

\bibitem{song2019dynamic}
Song, Y., Zhu, Y., Du, X.: Dynamic residual dense network for image denoising.
  Sensors  \textbf{19}(17), ~3809 (2019)

\bibitem{sun2018pwc}
Sun, D., Yang, X., Liu, M.Y., Kautz, J.: {PWC-N}et: {CNN}s for optical flow
  using pyramid, warping, and cost volume. In: CVPR. pp. 8934--8943 (2018)

\bibitem{sundaram2010dense}
Sundaram, N., Brox, T., Keutzer, K.: Dense point trajectories by
  {GPU}-accelerated large displacement optical flow. In: ECCV. pp. 438--451
  (2010)

\bibitem{tassano2019fastdvdnet}
Tassano, M., Delon, J., Veit, T.: Fast{DVD}net: Towards real-time video
  denoising without explicit motion estimation. arXiv preprint arXiv:1907.01361
   (2019)

\bibitem{tomasi1998bilateral}
Tomasi, C., Manduchi, R.: Bilateral filtering for gray and color images. In:
  ICCV (1998)

\bibitem{derfcollection}
Xiph.org: Derf's Test Media Collection ((accessed Nov 7, 2019)),
  \url{https://media.xiph.org/video/derf/}

\bibitem{xu2018real}
Xu, J., Li, H., Liang, Z., Zhang, D., Zhang, L.: Real-world noisy image
  denoising: A new benchmark. arXiv preprint arXiv:1804.02603  (2018)

\bibitem{xue2019video}
Xue, T., Chen, B., Wu, J., Wei, D., Freeman, W.T.: Video enhancement with
  task-oriented flow. IJCV  \textbf{127}(8),  1106--1125 (2019)

\bibitem{yue2019variational}
Yue, Z., Yong, H., Zhao, Q., Zhang, L., Meng, D.: Variational denoising
  network: Toward blind noise modeling and removal. In: NeurIPS (2019)

\bibitem{zhang2017beyond}
Zhang, K., Zuo, W., Chen, Y., Meng, D., Zhang, L.: Beyond a {G}aussian
  denoiser: Residual learning of deep {CNN} for image denoising. IEEE TIP
  (2017)

\bibitem{zhang2018ffdnet}
Zhang, K., Zuo, W., Zhang, L.: {FFDN}et: Toward a fast and flexible solution
  for {CNN}-based image denoising. IEEE TIP  \textbf{27}(9),  4608--4622 (2018)

\bibitem{zoran2011learning}
Zoran, D., Weiss, Y.: From learning models of natural image patches to whole
  image restoration. In: ICCV. pp. 479--486 (2011)

\end{thebibliography}
\end{document}